\documentclass[10pt,twocolumn,letterpaper]{article}

\usepackage[pagenumbers]{wacv} 

\newcommand{\arnela}[1]{{\color{black}#1}}   
\newcommand{\camred}[1]{{\color{black}#1}}

\definecolor{wacvblue}{rgb}{0.21,0.49,0.74}
\usepackage[pagebackref,breaklinks,colorlinks,allcolors=wacvblue]{hyperref}

\usepackage{comment}
\usepackage{tikz}
\usepackage{pgfplots}
\usepackage{booktabs}
\usepackage{multirow}

\title{VariaFlow: Uncertainty-Aware Flow Matching for \\Unsupervised Anomaly Detection}

\author{Arnela Hadzic$^{1}$, Martin Urschler$^{1,2}$, Cosmin I. Bercea$^{3}$\vspace{0.2cm}\\
{\small $^{1}$Institute for Medical Informatics, Statistics and Documentation, Medical University of Graz, Graz, Austria}\\[-2pt]
{\small $^{2}$Institute of Visual Computing, Graz University of Technology, Graz, Austria}\\[-2pt]
{\small $^{3}$Technical University of Munich, School of Medicine and Health, TUM University Hospital,}\\[-4pt]
{\small Klinikum rechts der Isar, Department of Diagnostic and Interventional Radiology, Munich, Germany}\\[-1pt]
{\tt\small arnela.hadzic@medunigraz.at}
}

\begin{document}
\maketitle
\begin{abstract}
Reconstruction-based anomaly detection no longer fails at reconstruction. Generative models produce high-quality pseudo-healthy reconstructions, and the remaining difficulty is the step of scoring residuals. 
The residual between a scan and its reconstruction has two indistinguishable sources, genuine pathology and the anatomical variation that separates one healthy brain from another. Scoring against a single reconstruction treats both as anomalous, inflating false positives where healthy anatomy varies most while small or subtle lesions are lost in the surrounding reconstruction error. We identify the missing quantity as the predictive variance, how much a healthy reconstruction can vary at each location. 
\camred{We introduce \emph{VariaFlow}, which estimates this variance from an ensemble of mask-guided restorations under an unconditional flow matching prior, each started from a different noise initialization, and which normalizes the residual by the resulting standard deviation. Few-step flow matching keeps the ensemble computationally affordable, and restoring at two spatial scales captures lesions of different size.} 
Across three brain MRI datasets spanning stroke and tumor, \emph{\mbox{VariaFlow}} is trained once on healthy anatomy and generalizes across all three, attaining the highest AUROC (0.94, 0.80, 0.78) \camred{on each} and the highest AUPRC on \camred{both cross-domain datasets}, exceeding the strongest baseline by 37\% and 50\%. \camred{Under domain shift}, the same baseline drops \camred{by up to} 52\% in AUPRC, whereas \emph{\mbox{VariaFlow}} stays \camred{more} stable and 
\camred{achieves the best small-lesion Dice on all three datasets.}

\end{abstract}
   
\section{Introduction}
\label{sec:intro}

\begin{figure*}[t]
    \centering
    \includegraphics[width=0.985\linewidth]{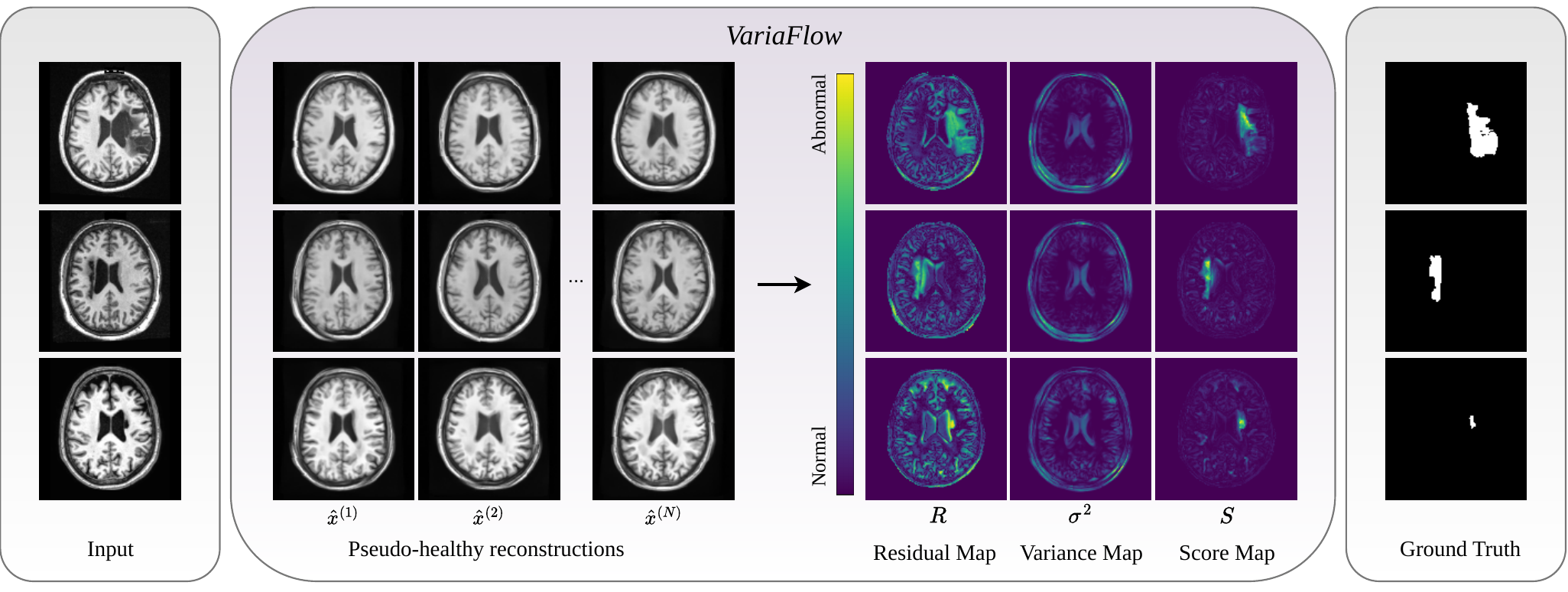}
    \caption{Given an input image, \emph{VariaFlow} generates an ensemble of pseudo-healthy reconstructions ($\hat{x}^{(1)}, \hat{x}^{(2)}, \dots, \hat{x}^{(N)}$). In anomalous regions, the model consistently generates healthy tissue, yielding a high residual ($R$) and low predictive variance ($\sigma^2$). Conversely, in variable healthy regions (e.g., ventricle boundaries), anatomical variability causes reconstructions to diverge, creating high predictive variance. Penalizing the residual with the variance suppresses false positives, isolating the pathology in the anomaly score map ($S$).}
    \label{fig:overview}
\end{figure*}

Brain anomaly detection localizes pathology in medical scans, such as tumors and stroke lesions. Supervised segmentation methods are effective~\cite{kamnitsas2016deepmedic,havaei2017brain,isensee2020nnu,sinha2020multi} but require pixel-wise lesion annotations and segment only the pathologies they were trained on.
Unsupervised methods instead learn the distribution of healthy anatomy and flag deviations from it, reconstructing an input as a pseudo-healthy image with a deep generative model and scoring the residual between the input and that reconstruction. 
\arnela{
Alternatively, feature-based methods detect anomalies by projecting images into a feature space via networks usually pre-trained on natural images \cite{roth2022towards,guo2023encoder}. 
Recent advancements extend this by extracting representations directly from large foundation models \cite{caron2021emerging,guo2025dinomaly,schulthess2025anomaly}.
However, the reliance on natural priors introduces a semantic gap, reducing their reliability in the medical domain. 
In contrast, reconstruction-based methods remain the dominant paradigm for brain MRI anomaly detection, as they yield intuitive pixel-level anomaly maps that are more suitable for segmentation tasks \cite{behrendt2026review}.}

\arnela{Among these reconstruction-based methods,} autoencoders~\cite{pawlowski2018unsupervised,zimmerer2018context,zimmerer2019unsupervised}
learn \arnela{the} distribution \arnela{of healthy anatomy} but output blurry reconstructions, which misses small or subtle pathology and inflates residuals from normal anatomical variation~\cite{bercea2023aes}.
While Generative Adversarial Networks (GANs) sharpen the reconstructions, they are unstable to train and prone to mode collapse~\cite{Schlegl2017unsupervised,schlegl2019f}. 
Diffusion models offer superior training stability and deliver high-fidelity reconstructions, but do so at a high cost of many iterative sampling steps~\cite{wyatt2022anoddpm,bercea2025melba,behrendt2024leveraging}. 
Flow matching models have gained in prominence to address this computational cost by learning straighter generative trajectories~\cite{lipman2022flow,liu2023flow}. The REFLECT method has recently adapted flow matching to generate pseudo-healthy reconstructions of anomalous scans~\cite{beizaee2025reflect}, but it does so in a compressed latent space which again leads to smoothing or even missing of subtle anatomical details or lesions. 

We hypothesize that a flow matching prior for efficient reconstruction-based anomaly detection requires the uncompressed input representation, as provided by recent flow priors for image restoration and inpainting~\cite{martin2025pnp, hadzic2026restora}. 
Moreover, we argue that the \arnela{standard practice} of aforementioned \arnela{reconstruction-based} methods, i.e.,  to score residuals against a single reconstruction, 
is insufficient. Healthy reconstruction is an ill-posed inverse problem. Once pathology is removed, many healthy anatomies can map to the same observation because individuating structure, such as cortical folding, is underdetermined by the input. The conditional distribution over healthy reconstructions is therefore diffuse, and a point estimate, the conditional mean or a single sample, discards its spread. False positive detections therefore concentrate exactly where anatomy is most variable.
\arnela{
Previous works have attempted to suppress false positives using restoration-based Maximum-A-Posteriori (MAP) estimation~\cite{chen2020unsupervised} or by modeling predictive uncertainty to calibrate anomaly scores~\cite{luo2025ura,gu2024revisiting}. 
However, such uncertainty-aware methods typically rely on heavily modified training pipelines and explicitly force ensemble disagreement~\cite{gu2024revisiting}, assuming that ensembles will primarily disagree on anomalous regions. 
We argue that for high-fidelity generative models, this assumption is reversed: ambiguous healthy regions inherently exhibit high variance across an ensemble, whereas anomalous inputs are consistently mapped to similar pseudo-healthy reconstructions.}

\arnela{Based on these insights,} we propose \emph{VariaFlow}, an uncertainty-aware anomaly detector that generates an ensemble of \camred{mask-guided restorations with an efficient flow matching prior.}
While pathologies and healthy anatomical variations both deviate from a single reconstruction, only the latter \camred{falls within the variation expected across the ensemble.}
Penalizing the median residual with the per-pixel variance effectively suppresses false positives in healthy regions. We extract these components at both a coarse and a fine masking scale, and fuse them via geometric mean before computing the final anomaly score, thus flagging a pixel only where both scales agree. We evaluate \emph{VariaFlow} on multiple brain MRI datasets spanning two pathologies, stroke and tumor, and three acquisition sources, BraTS, UCSF, and ATLAS.
Our main contributions are as follows:
\begin{itemize}
    \item We demonstrate that reconstruction-based anomaly detection greatly benefits from predictive variance, and show that scoring against a single reconstruction is what inflates false positives under anatomical variability.
    \item We introduce the first flow-based reconstruction ensemble for brain anomaly detection, \camred{where few-step flow matching keeps variance estimation computationally tractable.}
    \item We propose a variance-penalized anomaly score and a dual-scale geometric mean fusion that together suppress false positives from normal variation while preserving anomalies of varying size. 
    \item Across stroke and tumor lesions and three datasets, \emph{VariaFlow} improves detection and false positive suppression over diffusion- and flow-baselines, demonstrating superior robustness where the test cohort differs from training.
\end{itemize}

\section{Related Work}
\label{sec:related_work}
Unsupervised brain anomaly detection typically follows a corrupt-and-reconstruct paradigm~\cite{Schlegl2017unsupervised,pawlowski2018unsupervised,zimmerer2018context,zimmerer2019unsupervised,baur2021autoencoders,schlegl2019f,baur2020steganomaly}. Models learn the healthy anatomical distribution during training and restore pathological inputs to pseudo-healthy reconstructions at inference. Differences between input and reconstruction form the anomaly map.

Recent approaches predominantly utilize denoising diffusion models (DDPMs)~\cite{ho2020denoising,bercea2025melba}.  
Early diffusion methods relied on classifier guidance to translate diseased images to healthy ones~\cite{wolleb2022diffusion}, whereas recent unsupervised methods corrupt an input with noise and learn to reverse the process for a pseudo-healthy reconstruction.
However, standard diffusion introduces a sensitivity-precision trade-off, where excessive noise destroys fine healthy tissue details, while insufficient noise lets the model reconstruct the anomaly.
To address this, AnoDDPM~\cite{wyatt2022anoddpm} replaces standard Gaussian noise with Simplex noise, which resembles certain tumor appearances and improves degradation for those pathologies, although it misses anomalies where the noise structure differs from the training, e.g., structural anomalies such as fractures~\cite{bercea2024diffusion}. 
pDDPM~\cite{behrendt2024patched} applies diffusion in a patch-wise manner. However, training on patches is suboptimal to capture the global anatomical context.
Building on these foundations, methods such as AutoDDPM~\cite{bercea2023mask} and THOR~\cite{bercea2024diffusion} introduced more sophisticated guidance to refine the restorations. AutoDDPM combines automatic masking, stitching, and inpainting tasks, whereas THOR utilizes intermediate masks to reintegrate healthy tissue information during the reverse process. Inspired by the sequential refinement of diffusion models, IterMask$^{2}$~\cite{liang2024itermask} employs spatial masking with high-frequency guidance, iteratively shrinking the mask toward the anomaly. However, extracting high-frequency edge maps discards low-frequency information, forcing the model to infer tissue intensities from its training distribution, thus reducing robustness.

More recently, flow matching~\cite{lipman2022flow,liu2023flow} has been introduced as an alternative generative model to learn optimal transport paths between noise and data distributions, enabling more efficient inference compared to standard diffusion models. 
Specifically for image restoration, flow matching-based reconstruction priors provide substantial benefits in computational efficiency~\cite{hadzic2026restora} compared with state-of-the-art diffusion-based priors like RePaint~\cite{lugmayr2022repaint}.
To the best of our knowledge, REFLECT~\cite{beizaee2025reflect} is currently the only flow-based method for unsupervised brain anomaly detection. By simulating synthetic anomalies on healthy scans during training, it learns a direct correction trajectory within a compressed variational autoencoder (VAE) latent space. However, while achieving highly efficient inference, latent compression inherently smooths fine anatomical details, making the model prone to missing small lesions.

\section{Method}
\label{sec:method}

We propose \emph{VariaFlow}, an uncertainty-aware approach for unsupervised brain anomaly detection that leverages a generative flow matching prior trained on healthy anatomy. 
At inference, anomaly detection is formulated as a mask-guided image restoration task where the model projects the input image onto a healthy manifold.
Specifically, \emph{VariaFlow} generates an ensemble of pseudo-healthy reconstructions to yield both a median reconstruction and a predictive variance map. 
We combine these two components into an anomaly score map where the predictive variance acts as a pixel-wise penalty, effectively suppressing false positives in healthy variable regions (see \cref{fig:overview}). 
Furthermore, the reconstruction process is governed by a dual-scale masking strategy that enforces cross-scale structural agreement, allowing the model to robustly handle anomalies of diverse morphologies.

\subsection{Ensemble Restoration with Flow Prior}
\paragraph{Flow Matching.}
Flow matching~\cite{lipman2022flow,liu2023flow}  defines a continuous-time generative process by learning a time-dependent vector field $v_\theta(x, t)$. 
This field maps a simple base distribution, such as standard Gaussian noise $x_0 \sim \mathcal{N}(0,I)$, to a more complex data distribution via an ordinary differential equation (ODE):
\begin{equation*}
    \frac{dx}{dt} = v_\theta(x, t) .
\end{equation*} 
To approximate the target vector field $u_t(x_t)$, flow matching employs a simulation-free training objective:
\begin{equation*}
\mathcal{L}_{FM} = \mathbb{E}_{t \sim \mathcal{U}[0,1], x_0, x_1} \left[ \left\| v_\theta(x_t, t) - u_t(x_t) \right\|^2 \right].
\end{equation*}
A common and highly effective choice for the probability path $x_t$ is the linear interpolation between the initial noise $x_0$ and a target data sample $x_1$, formulated as $x_t = (1-t)x_0 + t x_1$. 
Consequently, the target velocity simplifies to $u_t = x_1 - x_0$.

During inference, a data sample can be generated by numerically integrating the learned ODE. 
For instance, using a standard Euler integration scheme with a step size $\Delta_t$, the state evolves iteratively:
\begin{equation*}
    x_{t+\Delta t} = x_t + \Delta_t  v_\theta(x_t, t) .
\end{equation*}

\paragraph{Pseudo-Healthy Reconstructions.}
Given a learned flow prior of healthy anatomy, we adopt a corrupt-and-restore paradigm to detect anomalies. 
We opt for spatial masking as our corruption mechanism, forcing the restoration to rely on the surrounding anatomical context when inpainting from the healthy data distribution.
Specifically, we process the input image $x_{in}$ using a dual-scale sliding window strategy defined by a set of distinct spatial scales $\mathcal{S} = \{s_l, s_g\}$, which represent a localized window $s_l$ and a larger contextual window $s_g$, respectively.

For each scale $s \in \mathcal{S}$, we apply a binary mask to the tissue within each window and employ Restora-Flow~\cite{hadzic2026restora} to restore the occluded regions. 
During each ODE integration step, the intermediate flow state $x_t$ is spatially fused with the known, unmasked regions of the input image, and a trajectory correction mechanism is applied to ensure alignment with the learned flow prior and the observed input~\cite{hadzic2026restora}. 
Finally, we aggregate the overlapping patches using a Hann window~\cite{pielawski2020introducing}, which ensures smooth transitions and yields a single pseudo-healthy reconstruction $\hat{x}_s$.

To construct an ensemble, we execute the sliding-window restoration process $N$ times per scale, using a distinct initial noise state $x_0$ for each individual run. This yields a scale-specific ensemble of $N$ independent reconstructions, denoted as $\mathcal{X}_s = \{\hat{x}_{s}^{(i)}\}_{i=1}^N$. 
As a result, structurally homogeneous regions yield consistent reconstructions across the $N$ samples, while regions of inherent structural ambiguity elevate the model's predictive variance.

\subsection{Variance-Aware Anomaly Scoring}
\paragraph{Scale-Specific Residuals and Variance.}
Given an ensemble $\mathcal{X}_s$ of pseudo-healthy reconstructions, we first compute the median reconstruction $\tilde{x}_s = \text{median}(\mathcal{X}_s)$ and derive the corresponding reconstruction residual 
\begin{equation}
\label{eq:median_residual}
R_s = |x_{in} - \tilde{x}_s|.
\end{equation}
Subsequently, we quantify the model's predictive uncertainty at that scale by computing the pixel-wise variance $\sigma_s^2$ across the $N$ reconstructions:
\begin{equation}
\sigma_s^2 = \frac{1}{N - 1} \sum_{i=1}^N (\hat{x}_{s}^{(i)} - \tilde{x}_s)^2.
\end{equation}

\paragraph{Geometric Dual-Scale Fusion.}
To aggregate information across localized and larger contextual windows, we apply \camred{geometric} mean fusion to the residuals and the variance maps. 
\camred{For the residuals, this acts as a spatial intersection, preserving only deviations that are consistent across both scales:}
\begin{equation}
\label{eq:median_residual_dual_scale}
R = \sqrt{R_{s_l} \cdot R_{s_g}}, \quad \sigma^2 = \sqrt{\sigma_{s_l}^2 \cdot \sigma_{s_g}^2} .
\end{equation}
While fusion using the arithmetic mean is also a valid option, we show in our ablation experiments that the proposed geometric mean fusion leads to slightly better suppression of false positives (see \Cref{subsec:ablations}).

\paragraph{Variance-Penalized Scoring.}
Finally, we compute the anomaly score $S$ as a variance-penalized residual. By \camred{normalizing} the reconstruction error by the ensemble standard deviation, we obtain
\begin{equation}
\label{eq:anomaly_score}
S = \frac{R}{\sigma + \gamma},
\end{equation}
where $\gamma$ acts as a regularization parameter. 
In this formulation, the denominator calibrates the score to distinguish healthy anatomical variations from pathology.
In variable healthy regions, high ensemble variance $\sigma^2$ attenuates the score, suppressing false positives. 
In contrast, anomalous regions yield a high residual $R$ but low variance $\sigma^2$, as the prior outputs consistent pseudo-healthy reconstructions. As a result, the denominator remains small and the resulting anomaly score highlights pathological deviations.

\section{Experimental Setup}
\label{sec:experimental_setup}
\textbf{Datasets.}
Our experimental setup utilizes four T1-weighted MRI datasets: IXI\footnote{\url{https://brain-development.org/ixi-dataset}} and ATLAS v2.0~\cite{liew2022large} for training and in-domain evaluation, alongside UCSF-PDGM~\cite{calabrese2022university} and BraTS2020~\cite{menze2014multimodal,bakas2017advancing,bakas2018identifying} for cross-domain testing. 
Following~\cite{bercea2024diffusion}, \arnela{we extract one mid-axial slice per patient}, normalize intensities to the 98th percentile, and pad and resize the images to a $128\times128$ resolution.
We train the model on healthy slices from ATLAS ($n=217$) and IXI ($n=465$), utilizing 58 healthy IXI slices for validation. 
In-domain testing is performed on 438 pathological \arnela{mid-axial} ATLAS slices with stroke lesions. 
Additionally, we evaluate the model's cross-domain generalization on 495 UCSF-PDGM and 369 BraTS pathological \arnela{mid-axial} slices with gliomas. 
As in~\cite{bercea2024diffusion}, we exclude 20 ATLAS \arnela{mid-axial} slices with prominent artifacts and any \arnela{mid-axial} slices with lesions smaller than 4 pixels from the evaluation. 
The remaining anomalies are stratified into small ($< 71$ pixels), medium, and large ($\ge 570$ pixels) lesions. 
\Cref{tab:image_distribution} provides the corresponding image distribution per lesion size category and dataset.

\begin{table}[t]
    \centering
    \caption{Number of images per lesion size category and dataset.}
    \label{tab:image_distribution}
    \small 
    \setlength{\tabcolsep}{12pt} 
    \begin{tabular}{lrrr}
        \toprule
         \textbf{Dataset} & \textbf{Small} & \textbf{Medium} & \textbf{Large} \\
         \midrule
         ATLAS     & 189 & 152 & 59 \\
         UCSF-PDGM & 30 & 126 & 257 \\
         BraTS     & 26  & 78  & 217 \\
         \midrule
         Total         & 245 & 356 & 533 \\
         \bottomrule
    \end{tabular}
\end{table}

\begin{table*}[t]
    \centering
    \caption{Quantitative results on the ATLAS, UCSF-PDGM, and BraTS datasets. \emph{VariaFlow} is compared to several state-of-the-art diffusion- and flow-based baselines. Bold and underline indicate the best and the second-best score per metric, respectively.}
    \label{tab:quantitative}
    \small 
    \setlength{\tabcolsep}{7pt} 

    \begin{tabular}{l ccc rrrr rr}
        \multicolumn{10}{c}{\textbf{ATLAS}} \\
        \toprule
        & & & & \multicolumn{4}{c}{\textbf{$\lceil$Dice$\rceil$ (\%) $\uparrow$}} & & \\
        \cmidrule(lr){5-8} 
        
        \textbf{Method} & \textbf{AUROC $\uparrow$} & \textbf{AUPRC $\uparrow$} & \textbf{P@S.5 (\%) $\uparrow$} & Average & Small & Medium & Large & \textbf{Sens (\%) $\uparrow$} & \textbf{Prec (\%) $\uparrow$} \\
        \midrule
        AutoDDPM~\cite{bercea2023mask}     & 0.7116 & 0.0878 & 7.30 & 13.86 & 3.25  & 15.29 & 25.87 & 27.02 & 9.32 \\
        THOR~\cite{bercea2024diffusion}    & 0.7139 & 0.1389 & 8.59 & 21.72 & 6.71 & 26.68 & 27.96 & 24.66 & 19.40 \\
        IterMask²~\cite{liang2024itermask} & 0.8582 & 0.4965 & 48.40 & 50.07 & 7.79  & 44.52 & 59.86 & 45.00 & \textbf{56.43} \\
        REFLECT~\cite{beizaee2025reflect}  & \underline{0.9378} & \textbf{0.5812} & \textbf{61.14} & \textbf{55.86} & \underline{9.42} & \underline{46.86} & \textbf{66.78} & \underline{56.62} & \underline{55.13} \\
        \midrule
        \emph{VariaFlow}                      & \textbf{0.9417} & \underline{0.5262} & \underline{57.19} & \underline{54.86} & \textbf{18.88} & \textbf{51.76} & \underline{65.05} & \textbf{58.94} & 51.31 \\
        \bottomrule
    \end{tabular}

    \vspace{0.2cm} 
    
    \begin{tabular}{l ccc rrrr rr}
        \multicolumn{10}{c}{\textbf{UCSF-PDGM}} \\
        \toprule
        & & & & \multicolumn{4}{c}{\textbf{$\lceil$Dice$\rceil$ (\%) $\uparrow$}} & & \\
        \cmidrule(lr){5-8} 
        
        \textbf{Method} & \textbf{AUROC $\uparrow$} & \textbf{AUPRC $\uparrow$} & \textbf{P@S.5 (\%) $\uparrow$} & Average & Small & Medium & Large & \textbf{Sens (\%) $\uparrow$} & \textbf{Prec (\%) $\uparrow$} \\
        \midrule
        AutoDDPM~\cite{bercea2023mask}     & 0.6566 & 0.2452 & 23.18 & 31.52 & 1.86 & \underline{17.12} & 38.86 & 51.59 & 22.33 \\
        THOR~\cite{bercea2024diffusion}    & 0.6512 & 0.2229 & 22.13 & 30.54 & \underline{2.10} & 16.31 & 36.54 & 44.44 & 23.27 \\
        IterMask²~\cite{liang2024itermask} & 0.6159 & 0.2016 & 17.96 & 26.71 & 0.93 & 13.60 & 31.73 & \textbf{54.73} & 17.66 \\
        REFLECT~\cite{beizaee2025reflect}  & \underline{0.7235} & \underline{0.3079} & \underline{27.48} & \underline{35.18} & 0.74 & 16.10 & \underline{41.95} & 50.72 & \underline{26.93} \\
        \midrule
        \emph{VariaFlow}                       & \textbf{0.7952} & \textbf{0.4222} & \textbf{40.15} &  \textbf{44.16} & \textbf{2.44} & \textbf{27.92} & \textbf{49.28} & \underline{52.11} & \textbf{38.31} \\
        \bottomrule
    \end{tabular}

    \vspace{0.2cm} 
    
    \begin{tabular}{l ccc rrrr rr}
        \multicolumn{10}{c}{\textbf{BraTS}} \\
        \toprule
        & & & & \multicolumn{4}{c}{\textbf{$\lceil$Dice$\rceil$ (\%) $\uparrow$}} & & \\
        \cmidrule(lr){5-8} 
        
        \textbf{Method} & \textbf{AUROC $\uparrow$} & \textbf{AUPRC $\uparrow$} & \textbf{P@S.5 (\%) $\uparrow$} & Average & Small & Medium & Large & \textbf{Sens (\%) $\uparrow$} & \textbf{Prec (\%) $\uparrow$} \\
        \midrule
        AutoDDPM~\cite{bercea2023mask}     & 0.6108 & 0.2276 & 21.19 & 29.60 & \underline{2.12} & \underline{14.99} & 35.82 & 52.71 & 20.58 \\
        THOR~\cite{bercea2024diffusion}    & 0.6218 & 0.2198 & 21.38 & 29.65 & 1.97 & 14.27 & 35.23 & 48.33 & 21.38 \\
        IterMask²~\cite{liang2024itermask} & 0.5821 & 0.1904 & 17.88 & 26.93 & 1.39 & 13.12 & 31.59 & \textbf{58.71} & 17.47 \\
        REFLECT~\cite{beizaee2025reflect}  & \underline{0.6828} & \underline{0.2779} & \underline{24.97} &\underline{33.16} & 0.64 & 14.95 & \underline{38.80} & \underline{55.66} & \underline{23.62} \\
        \midrule
        \emph{VariaFlow}               & \textbf{0.7800} & \textbf{0.4162} & \textbf{39.45} & \textbf{43.65} & \textbf{2.57} & \textbf{23.73} & \textbf{48.59} & 52.95 & \textbf{37.13} \\
        \bottomrule
    \end{tabular}
\end{table*}

\noindent
\textbf{Evaluation Metrics and Baselines.}
Following related work~\cite{bercea2024diffusion,liang2024itermask,beizaee2025reflect}, we assess binary segmentation performance by first determining the optimal threshold based on the maximum Dice coefficient ($\lceil$Dice$\rceil$) across the range $[0, 1]$. We then apply this threshold to compute Sensitivity (Sens), Precision (Prec), and Dice scores for different lesion categories (Small, Medium, Large). 
Additionally, we evaluate continuous anomaly maps using threshold-agnostic metrics. We use AUROC to measure how well a model distinguishes anomalies from healthy anatomy, as well as AUPRC to reflect false positive rates independently of the predominant healthy background and to address shortcomings of AUROC in the presence of imbalanced datasets.
Additionally, we measure false positive suppression capabilities using Precision at 50\% Sensitivity (P@S.5) to quantify precision at a fixed 50\% operating point.
We compare \emph{VariaFlow} to several recent state-of-the-art baselines, including diffusion-based AutoDDPM~\cite{bercea2023mask} with iterative resampling, THOR~\cite{bercea2024diffusion} utilizing implicit guidance, IterMask$^{2}$~\cite{liang2024itermask} featuring iterative masking, and flow-based 
REFLECT~\cite{beizaee2025reflect}, \arnela{utilizing their official implementations.}
\arnela{All compared methods were trained from scratch on the same datasets.}

\noindent
\textbf{Implementation Details.}
Our unconditional flow matching prior model~\cite{lipman2022flow} uses a U-Net~\cite{Ronneberger2015-ih} network architecture and was trained for $1500$ epochs using the AdamW optimizer~\cite{loshchilov2019adamw}, a learning rate of $2\times10^{-5}$, and a batch size of 16. All images were normalized to $[-1, 1]$. 
At inference, a dual-scale sliding window with $50\%$ overlap extracts 49 local windows at the $32\times32$ scale ($s_l=32$), and 9 windows at the $64\times64$ scale ($s_g=64$). Each window is independently inpainted using 8 ODE steps. Restored windows are aggregated per scale.
The regularization factor $\gamma$ in \cref{eq:anomaly_score} was empirically set to $0.025$, \arnela{as it yielded the highest AUPRC in our initial experiments.} 

\section{Results}
\label{sec:results}

\begin{figure*}[t]
    \centering
    \includegraphics[width=0.99\linewidth]{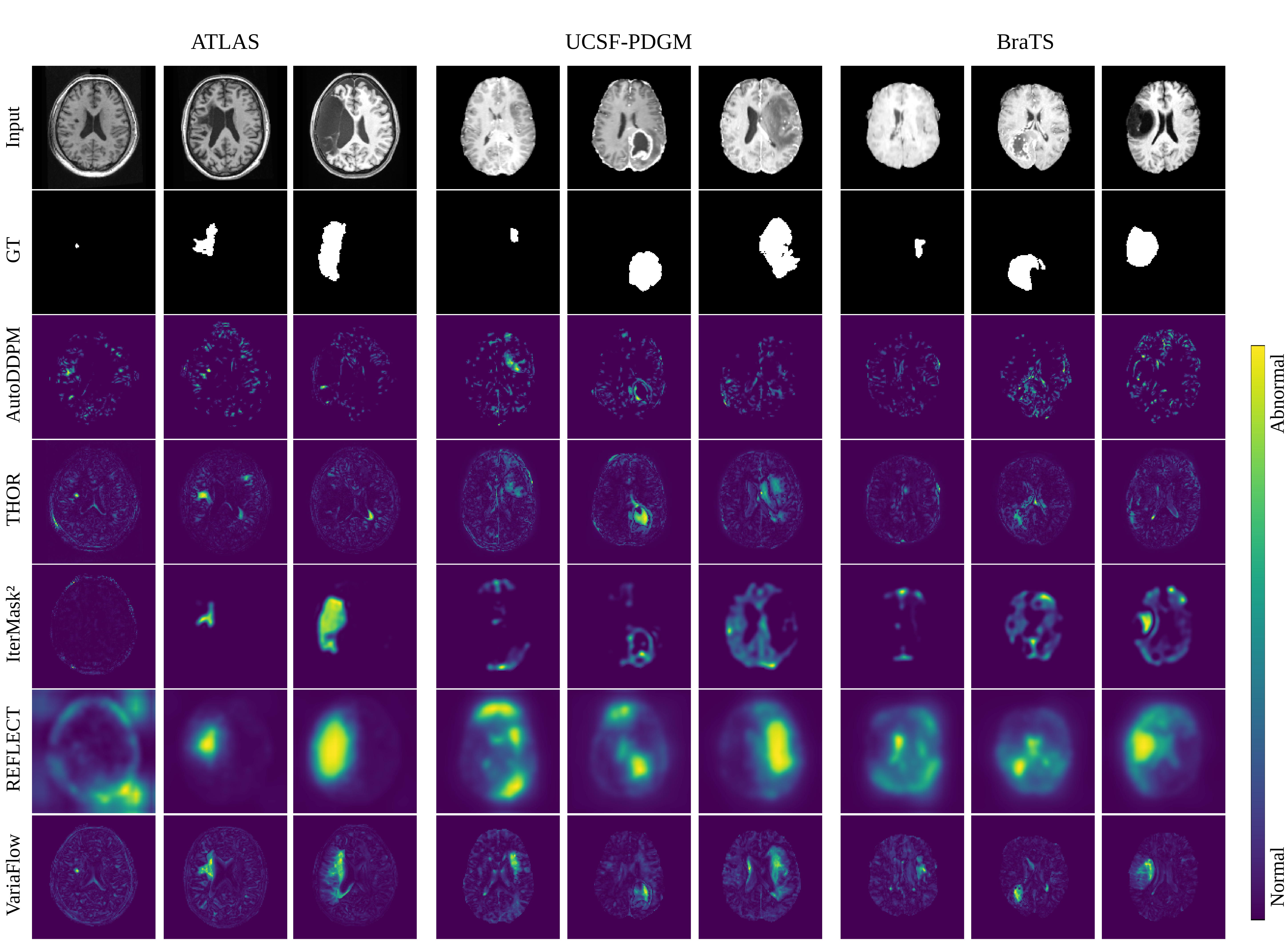}
    \caption{Qualitative anomaly detection results on ATLAS, UCSF-PDGM, and BraTS. Rows 1-2 show the input images and ground truth masks. Rows 3-6 visualize the anomaly maps generated by the baseline methods, and Row 7 shows the maps generated by \emph{VariaFlow}, demonstrating both superior detection and false positive suppression.
    }
    \label{fig:qualitative}
\end{figure*}

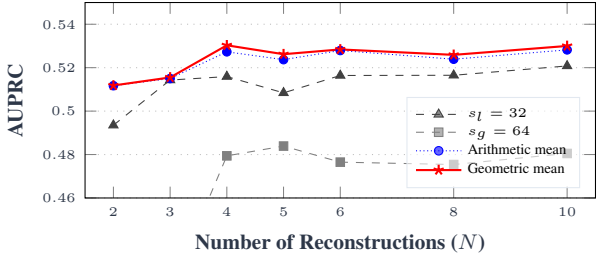
\begin{figure}[t]
    \definecolor{charcoal}{HTML}{2D3748}
\definecolor{bordergray}{HTML}{E2E8F0}
\pgfplotsset{compat=1.18}
\begin{tikzpicture}
    \begin{axis}[
        width=\linewidth,
        height=0.5\linewidth,
        xlabel={\textbf{Number of Reconstructions ($N$)}},
        ylabel={\textbf{AUPRC}},
        xlabel style={text=charcoal,font=\footnotesize},
        ylabel style={text=charcoal,font=\footnotesize,yshift=0.5pt},
        xmin=1.5, xmax=10.5,
        ymin=0.46, ymax=0.55,
        xtick={2, 3, 4, 5, 6, 8, 10},
        ytick={0.46, 0.48, 0.50, 0.52, 0.54},
        tick label style={font=\tiny\sffamily, text=charcoal},
        legend pos=south east,
        legend style={
        at={(0.97,0.05)}, 
        font=\fontsize{5.5pt}{2pt}\selectfont,
        fill=white,  
        fill opacity=0.6, 
        text opacity=1,
        draw=bordergray, 
        cells={anchor=west}, 
        row sep=-2pt
        },
        ymajorgrids=true,
        grid style=dotted,
        axis background/.style={fill=white}
    ]
    \addplot[
        color=darkgray, 
        dashed, 
        mark=triangle*, 
        mark options={solid, scale=1}
    ] coordinates {
        (2, 0.4935)
        (3, 0.5143)
        (4, 0.5159)
        (5, 0.5084)
        (6, 0.5164)
        (8, 0.5165)
        (10, 0.5208)
    };
    \addlegendentry{$s_l=32$}
    \addplot[
        color=gray, 
        dashed, 
        mark=square*, 
        mark options={solid, scale=0.8}
    ] coordinates {
        (2, 0.4155)
        (3, 0.4176)
        (4, 0.4794)
        (5, 0.4839)
        (6, 0.4765)
        (8, 0.4754)
        (10, 0.4805)
    };
    \addlegendentry{$s_g=64$}
    \addplot[
        color=blue, 
        densely dotted, 
        mark=*, 
        mark options={solid, scale=0.8}
    ] coordinates {
        (2, 0.5117)
        (3, 0.5150)
        (4, 0.5273)
        (5, 0.5237)
        (6, 0.5279)
        (8, 0.5239)
        (10, 0.5282)
    };
    \addlegendentry{Arithmetic mean}
    \addplot[
        color=red, 
        thick, 
        solid, 
        mark=star, 
        mark options={solid, scale=1}
    ] coordinates {
        (2, 0.5118)
        (3, 0.5155)
        (4, 0.5303)
        (5, 0.5262)
        (6, 0.5284)
        (8, 0.5259)
        (10, 0.5300)
    };
    \addlegendentry{Geometric mean}
    \end{axis}
\end{tikzpicture}  
    \caption{Comparison of window sizes ($s_l=32, s_g=64$) and fusion strategies, with geometric mean consistently yielding the highest AUPRC.}
    \label{fig:ablation1}
\end{figure}

\Cref{tab:quantitative} shows the quantitative results for \emph{VariaFlow} against state-of-the-art baselines across the ATLAS, UCSF-PDGM, and BraTS datasets. \emph{VariaFlow} achieves superior robustness and generalization, outperforming existing diffusion and flow-based methods on unseen domains while remaining highly competitive in-domain. Qualitative results in \cref{fig:qualitative} visually validate these findings, while \cref{fig:ablation1} and \cref{fig:ablation2} analyze the impact of our dual-scale strategy and of the ensemble size, respectively. 
\Cref{tab:ablation3} demonstrates the effectiveness of predictive variance in suppressing false positives.
Lastly, \cref{fig:failure_cases} illustrates two scenarios where our approach still faces challenges in anomaly detection.

\subsection{Quantitative Results} 
On the in-domain ATLAS dataset, \emph{VariaFlow} achieves the new state-of-the-art AUROC score for anomaly detection and highly competitive AUPRC as well as Average $\lceil$Dice$\rceil$ scores for false positive suppression and segmentation capability, respectively.
While REFLECT achieves higher overall AUPRC and \camred{a marginally higher} Average $\lceil$Dice$\rceil$, its reliance on latent-space compression likely limits its ability to resolve fine details. 
Although detecting small-sized lesions remains inherently challenging for all unsupervised methods, \emph{VariaFlow} is  more robust at these scales. 
Specifically, \emph{VariaFlow} doubles the performance of REFLECT on small anomalies (18.88\% vs. 9.42\% $\lceil$Dice$\rceil$) and notably improves upon medium-sized anomalies (51.76\% vs. 46.86\%). 
\camred{This is consistent with REFLECT's design, where compressing the input into a VAE latent space before applying flow matching likely smooths over fine anatomical details and subtle lesions.}
Conversely, \emph{VariaFlow}'s spatially conditioned restoration 
avoids the information loss \camred{introduced by} the latent-space strategy, thus better preserving the high-frequency anatomical details necessary for detecting subtle pathologies.

In the more challenging cross-domain evaluations using UCSF-PDGM and BraTS, the diffusion-based baselines AutoDDPM and THOR show higher Average $\lceil$Dice$\rceil$ scores than on ATLAS, likely due to the presence of more prominent tumors in these datasets that are easier to detect. 
While domain shift challenges all unsupervised methods, top-performing in-domain baselines IterMask$^{2}$ and REFLECT suffer a more severe performance drop in most metrics.
For instance, REFLECT's Average $\lceil$Dice$\rceil$ drops from 55.86\% on ATLAS to 35.18\% on UCSF-PDGM and from 0.9378 to 0.6828 AUROC on BraTS. 
IterMask$^{2}$ shows a similar trend, as its domain-specific high-frequency guidance seems to overfit to the training domain's intensities. 
In contrast, \emph{VariaFlow} remains more stable when generalizing to both domains, achieving state-of-the-art results across nearly all metrics.
Notably, on UCSF-PDGM \emph{VariaFlow} improves AUROC by 9.9\% over the next best baseline REFLECT and achieves a new state-of-the-art Average $\lceil$Dice$\rceil$ of 44.16\%.
Moreover, on BraTS, \emph{VariaFlow} improves 14.2\% upon the second best method in AUROC and achieves an Average $\lceil$Dice$\rceil$ of 43.65\%, an increment of 10.49\%, thus demonstrating the superior detection performance and robustness compared to baseline approaches.

\begin{table}[b]
    \centering
    \caption{Comparison of different scoring formulations. Median residual alone as proposed in \cref{eq:median_residual_dual_scale}. Mean residual is a variant that uses the mean reconstruction instead in \cref{eq:median_residual}. Based on median residual, our proposed $(S)$ additionally penalizes variance.}
    \label{tab:ablation3}
    \small 
    \begin{tabular}{lccc}
        \toprule
         \textbf{Scoring} & \textbf{AUROC} $\uparrow$ & \textbf{AUPRC} $\uparrow$ & \textbf{$\lceil$Dice$\rceil$ (\%)} $\uparrow$ \\
         \midrule
         Mean residual    & 0.8463 & 0.3935 & 43.11 \\
         Median residual & 0.8571 & 0.4008 & 43.91 \\
         \midrule
         Proposed $(S)$              & \textbf{0.9417} & \textbf{0.5262} & \textbf{54.86} \\
         \bottomrule
    \end{tabular}
\end{table}

Beyond detection performance, our cross-domain generalization results on UCSF-\camred{PDGM} and BraTS also demonstrate \textit{VariaFlow's} major improvement in false positive suppression. For all three false positive related metrics (AUPRC, P@S.5, Prec), improvements upon the second best method REFLECT range between 37\% and \camred{58}\%.
This underlines that our strategy of uncertainty-aware dual-scale flow-based reconstruction can achieve accurate anomaly detection without the trade-off of excessive false positives.

\subsection{Qualitative Results}

Anomaly maps generated by diffusion-based methods AutoDDPM and THOR (see \cref{fig:qualitative}, Rows 3-4) show reasonable overall detection performance, however, at the price of generating a large number of false positive detections. IterMask$^{2}$ (Row 5) also visually demonstrates an improved detection and false positive suppression, however, it does so only in the in-domain scenario due to its domain-specific high frequency guidance. REFLECT (Row 6) provides better detection than the other baselines, 
\camred{but its latent-space compression likely limits the spatial precision of its anomaly maps, which further hinders the detection of small and subtle lesions. In the cross-domain setting, its maps additionally exhibit many false positives.}
In contrast, \emph{VariaFlow} \camred{yields well-localized anomaly maps across lesion sizes, with the largest gains in false positive suppression in the challenging cross-domain setting.} 
The importance of predictive variance to derive the anomaly score is also demonstrated in \cref{fig:overview} via additional ATLAS examples, showing improved false positive suppression.

\begin{figure}[t]
    \centering
    \includegraphics[width=\linewidth]{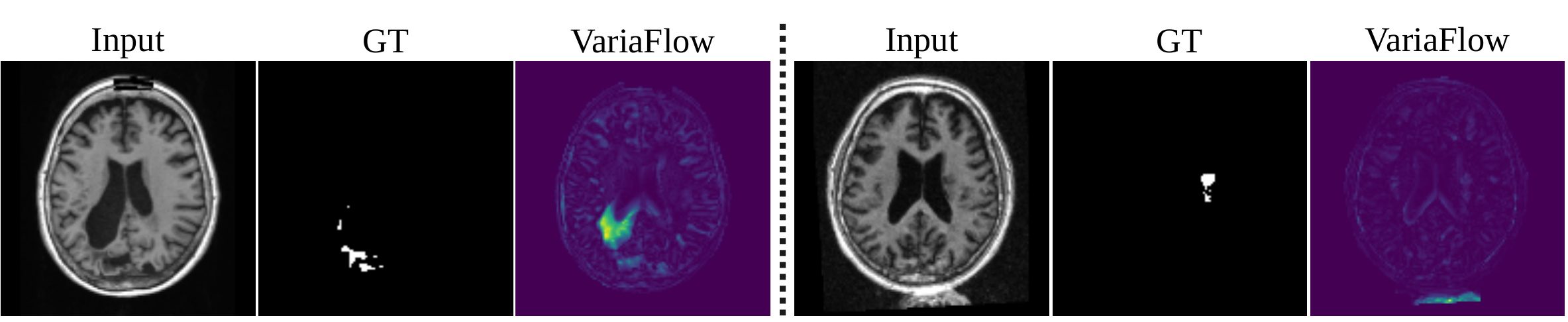}
    \caption{Examples of diverse anatomical and structural deviations. Left: The model detects an enlarged ventricle, dominating the lesion attached to it. Right: The model misses a lesion resembling cortical folding while detecting a prominent imaging artifact.}
    \label{fig:failure_cases}
\end{figure}

While \emph{VariaFlow} \camred{achieves strong detection} and false positive suppression, we have identified a few cases in our testing data where our approach still faces challenges.
The left part of \cref{fig:failure_cases} illustrates a case with low performance in Dice score. \emph{VariaFlow} flags an enlarged ventricle adjacent to the annotated infarct, reducing Dice against the ground truth. The enlargement is consistent with ex vacuo dilatation, the expansion of the ventricle into space left by infarcted tissue, and is a genuine downstream effect of the stroke. ATLAS annotates only the infarct core, so this secondary change is absent from the ground truth despite being clinically real. The change is also diagnostic. Chronic infarction destroys tissue and the ventricle expands into the resulting void, whereas a mass such as a tumor compresses and displaces it. The direction of ventricular deformation reflects the underlying process, volume loss versus mass effect, and bears directly on distinguishing chronic stroke from neoplasm. The right part of \cref{fig:failure_cases} shows a subtle lesion that closely resembles normal surrounding tissue, which likely explains the missed detection. \emph{VariaFlow} additionally responds to an imaging artifact at the skull base, counted as a false positive, though this too is a true deviation from healthy anatomy.

\begin{figure}[t]
    \centering
    \definecolor{charcoal}{HTML}{2D3748}
\definecolor{bordergray}{HTML}{E2E8F0}
\definecolor{lightblue}{HTML}{8CC5E3}
\definecolor{medblue}{HTML}{3594CC}
\definecolor{lightorange}{HTML}{F0B077}
\definecolor{medorange}{HTML}{EA801C}
\pgfplotsset{compat=1.18}
\begin{tikzpicture}
  \begin{axis} [
    scale only axis,
    width=0.8\linewidth,  
    height=0.45\linewidth,
    grid=major,
    grid style={dotted, gray!60},
    xtick={1, 2, 3, 4, 5, 6, 8, 10},
    xmin=0.5, xmax=10.5,
    ymin=0.30, ymax=1.00,
    ytick={0.3, 0.4, 0.5, 0.6, 0.7, 0.8, 0.9, 1.0},
    xlabel={\textbf{Number of Reconstructions ($N$)}},
    ylabel={\textbf{AUROC / AUPRC}},
    xlabel style={text=charcoal,font=\footnotesize},
    ylabel style={text=charcoal,font=\footnotesize,yshift=0.5pt},
    tick label style={font=\tiny\sffamily, text=charcoal},
    axis y line*=left,
    axis x line*=bottom,
    legend style={
      at={(0.97,0.38)}, 
      anchor=south east, 
      draw=bordergray, 
      fill=white, 
      font=\fontsize{5.5pt}{2pt}\selectfont,
      cells={anchor=west},
      row sep=-2pt
    },
    legend cell align={left}
  ]
    \addplot[color=lightblue, mark=*, mark options={solid}, thick, mark size=1pt] coordinates {
      (1, 0.8376) (2, 0.9262) (3, 0.9375) (4, 0.9418) 
      (5, 0.9417) (6, 0.9410) (8, 0.9403) (10, 0.9418)
    };
    \addlegendentry{AUROC}
    \addplot[color=medblue, mark=square*, mark options={solid}, thick, dashed, mark size=1pt] coordinates {
      (1, 0.3545) (2, 0.5118) (3, 0.5155) (4, 0.5303) 
      (5, 0.5262) (6, 0.5284) (8, 0.5259) (10, 0.5300)
    };
    \addlegendentry{AUPRC}
    \addlegendimage{color=medorange, mark=triangle*, mark options={solid}, thick, dashdotdotted, mark size=1.5pt}
    \addlegendentry{Average $\lceil$Dice$\rceil$}
    \addlegendimage{color=lightorange, mark=star, mark options={solid}, thick, dashdotdotted, mark size=2pt}
    \addlegendentry{Average $\lceil$Dice$\rceil$ (small)}
  \end{axis}
  \begin{axis}[
    scale only axis,
    width=0.8\linewidth,  
    height=0.45\linewidth,
    xmin=0.5, xmax=10.5,
    ymin=0, ymax=70,
    ytick={0, 10, 20, 30, 40, 50, 60, 70},
    hide x axis, 
    axis y line*=right, 
    ylabel={\textbf{Average $\lceil$Dice$\rceil$ (\%)}},
    ylabel style={text=medorange, yshift=0.5pt,font=\footnotesize},
    yticklabel style={text=medorange},
    tick label style={font=\tiny\sffamily}
  ]
    \addplot[color=medorange, mark=triangle*, mark options={solid}, mark options={solid}, thick, dashdotdotted, mark size=1.5pt] coordinates {
      (1, 39.59) (2, 54.31) (3, 54.07) (4, 55.11) 
      (5, 54.86) (6, 54.90) (8, 54.41) (10, 54.86)
    };
    \addplot[color=lightorange, mark=star, mark options={solid}, mark options={solid}, thick, dashdotdotted, mark size=2pt] coordinates {
      (1, 4.07) (2, 13.95) (3, 16.50) (4, 17.14) 
      (5, 18.88) (6, 19.42) (8, 19.59) (10, 19.95)
    };
  \end{axis}
\end{tikzpicture}%
    \caption{Impact of the ensemble size $N$ on detection performance. Metrics stabilize at $N=5$.}
    \label{fig:ablation2}
\end{figure}
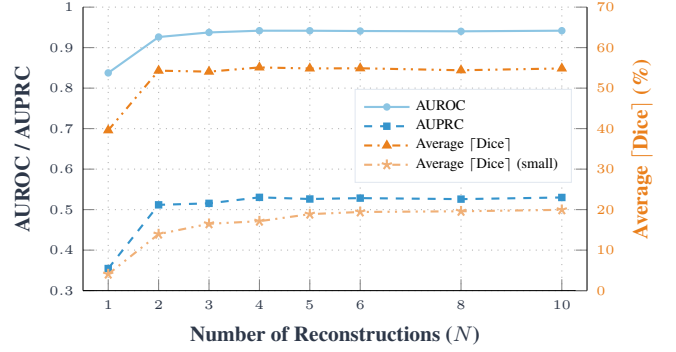

\begin{table*}[t]
    \centering
    \caption{Computational and architectural comparison between methods.
    }
    \label{tab:computational_analysis}
    \small 
    \setlength{\tabcolsep}{7pt} 
    \begin{tabular}{llrrrr}
        \toprule
        \textbf{Method} & \textbf{Backbone Architecture} & \textbf{Steps} & \textbf{Time (s)} & \textbf{VRAM (GB)} & \textbf{Params (M)} \\
        \midrule
        THOR \cite{bercea2024diffusion} & DDPM U-Net & 350 & 3.58 & 0.52 & $\sim 19$ \\
        REFLECT \cite{beizaee2025reflect} & LDM U-Net + VAE & 5 & 0.12 & 1.67 & $\sim 246$ \\
        \midrule
        \emph{VariaFlow} (single patch) & \multirow{5}{*}{Restora-Flow U-Net \cite{hadzic2026restora}} & \multirow{5}{*}{8} & 0.32 & 0.40 & \multirow{5}{*}{$\sim 62$} \\
        \emph{VariaFlow} (global scale, $N=1$) & & & 0.64 & 0.82 & \\
        \emph{VariaFlow} (local scale, $N=1$) & & & 2.92 & 3.34 & \\
        \emph{VariaFlow} (dual-scale, $N=1$) & & & \camred{3.56} & 3.34 & \\
        \bottomrule
    \end{tabular}
\end{table*}

\subsection{Computational Aspects}
\arnela{
\Cref{tab:computational_analysis} provides a runtime analysis across diffusion-based THOR~\cite{bercea2024diffusion}, flow-based REFLECT~\cite{beizaee2025reflect}, and our method, measured on a single NVIDIA A100 GPU (40 GB).
}

\arnela{
REFLECT is the fastest approach, requiring 0.12\,s to generate a single reconstruction, as it operates within a heavily compressed latent space. 
While computationally efficient, REFLECT exhibits reduced preservation of high-frequency anatomical details and weaker cross-domain generalization in our experiments (\cref{fig:qualitative,tab:quantitative}).
}

\arnela{
For pixel-space generation, THOR and \emph{VariaFlow} have comparable wall-clock runtimes of approximately 3.5\,s per reconstruction, despite relying on fundamentally different generation strategies.
THOR utilizes a DDPM U-Net ($\sim$19\,M parameters) and requires 350 sequential reverse diffusion steps, whereas \emph{VariaFlow} achieves this runtime by parallelizing restorations of 58 patches using a Restora-Flow U-Net ($\sim$62\,M parameters) and an efficient 8-step flow matching prior.
}

\arnela{
Specifically, generating a single full-image reconstruction (dual-scale, $N=1$) with our method involves parallelizing 49 local-scale ($s_l=32$, 2.92\,s) and 9 global-scale ($s_g=64$, 0.64\,s) windows. 
Since Restora-Flow~\cite{hadzic2026restora} processes a fixed-size full-resolution tensor and represents each target window only through its binary mask, the cost per window remains approximately constant at 0.32\,s (8 ODE steps), regardless of the mask size. 
Parallelizing all 58 windows on the GPU compresses what would sequentially take over 18\,s into just \camred{3.56}\,s.
Importantly, the dual-scale \emph{VariaFlow} configuration with $N=1$ already outperforms THOR (\cref{fig:ablation2,tab:quantitative}) at a comparable runtime.
}

Our main configuration uses an ensemble of $N=5$ independent reconstructions. Executed sequentially, this \camred{takes 17.8\,s with a peak memory of 3.34\,GB, while the subsequent fusion and scoring add only negligible overhead.}

\subsection{Ablations and Limitations} 
\label{subsec:ablations}

\textbf{Ablation of window scales.} 
\cref{fig:ablation1} illustrates that fusing dual-scale outputs consistently improves the overall performance compared to single-scale baselines, with the geometric mean from \cref{eq:median_residual_dual_scale} slightly outperforming the fusion via the arithmetic mean.
Because it requires strict spatial agreement across both scales, the geometric mean fusion penalizes single-scale noise, yielding the highest AUPRC scores.

\noindent
\textbf{Sensitivity analysis of ensemble size.} 
Using the dual-scale output fusion via the geometric mean, \cref{fig:ablation2} demonstrates the impact of varying the number of reconstructions $N$ in the ensemble. At $N=1$, the framework relies on a single fused estimate \arnela{($\sigma=0$)}.
Constructing a generative ensemble at $N=2$ notably improves detection performance, increasing Average $\lceil$Dice$\rceil$ from approximately $40\%$ to $54\%$ and AUPRC from $0.35$ to $0.51$. 
While overall performance continues to improve \arnela{only marginally} with larger values of $N$, \arnela{relying on only two reconstructions can introduce adverse effects, such as both randomly aligning on a similar anatomy and failing to suppress a false positive.}
\arnela{Further, the detection of small lesions continues to benefit from a more robust variance estimate.}
\camred{All metrics stabilize at $N = 5$, which we select as our optimal setup, since scaling to $N = 10$ doubles the inference time for negligible gains.}

\noindent
\textbf{Effect of predictive variance.}
\Cref{tab:ablation3} shows the impact of predictive variance by comparing different scoring functions at a fixed ensemble size ($N=5$) on ATLAS. We evaluated the median residual $(R)$ from \cref{eq:median_residual_dual_scale} on its own as a scoring function, as well as a variant where the mean reconstruction is used instead of the median in \cref{eq:median_residual}. Experimentally, the median residual provides slightly improved detection performance, however, relying solely on the residual still fails to account for natural anatomical deviations. By penalizing the median residual with the predictive variance, our proposed score $S$ achieves a substantial increase in detection performance (AUROC, $\lceil$Dice$\rceil$), while at the same time increasing AUPRC by 31\% from 0.4008 to 0.5262, thus successfully suppressing false positives in variable healthy regions.

\noindent
\textbf{Limitations.}
\camred{To enable a controlled comparison with prior work, we evaluate on 2D mid-axial slices only and exclude lesions smaller than 4 pixels. While the ensemble-based scoring itself extends directly to 3D, the number of sliding windows grows cubically, so a volumetric extension might require more efficient sub-volume restoration.}

\section{Conclusion}
\label{sec:conclusion}
We have introduced \emph{VariaFlow}, an uncertainty-aware framework for unsupervised brain anomaly detection. \emph{VariaFlow} addresses the primary limitation of single-reconstruction methods, where healthy anatomical variability is frequently misidentified as pathology. By efficiently generating a flow matching-based ensemble, our method utilizes a variance-penalized anomaly score and a dual-scale strategy to effectively detect anomalies and suppress false positives across varying lesion sizes. Experimental results on in-domain and cross-domain data demonstrate that \emph{VariaFlow} outperforms existing diffusion- and flow-based baselines, maintaining robust performance across multiple brain MRI datasets. 

\section*{Acknowledgements}
This research was funded in whole or in part by the Austrian Science Fund (FWF) 10.55776/PAT1748423. The computational results have been achieved using the Austrian Scientific Computing (ASC) infrastructure.

{
    \small
    \bibliographystyle{ieeenat_fullname}
    \bibliography{main}

@String(CVPR= {IEEE Conf. Comput. Vis. Pattern Recog.})

@String(ICCV= {Int. Conf. Comput. Vis.})

@String(ICLR = {Int. Conf. Learn. Represent.})

@String(CVPR  = {CVPR})

@String(ICCV  = {ICCV})

@String(ICLR  = {ICLR})

@inproceedings{kamnitsas2016deepmedic,
  title={{DeepMedic for brain tumor segmentation}},
  author={Kamnitsas, Konstantinos and Ferrante, Enzo and Parisot, Sarah and Ledig, Christian and Nori, Aditya V and Criminisi, Antonio and Rueckert, Daniel and Glocker, Ben},
  booktitle={International Workshop on Brainlesion: Glioma, Multiple Sclerosis, Stroke and Traumatic Brain Injuries},
  pages={138--149},
  year={2016}
}

@article{havaei2017brain,
  title={Brain tumor segmentation with deep neural networks},
  author={Havaei, Mohammad and Davy, Axel and Warde-Farley, David and Biard, Antoine and Courville, Aaron and Bengio, Yoshua and Pal, Chris and Jodoin, Pierre-Marc and Larochelle, Hugo},
  journal={Medical Image Analysis},
  volume={35},
  pages={18--31},
  year={2017}
}

@inproceedings{isensee2020nnu,
  title={{nnU-Net for brain tumor segmentation}},
  author={Isensee, Fabian and J{\"a}ger, Paul F and Full, Peter M and Vollmuth, Philipp and Maier-Hein, Klaus H},
  booktitle={International Workshop on Brainlesion: Glioma, Multiple Sclerosis, Stroke and Traumatic Brain Injuries},
  pages={118--132},
  year={2020}
}

@article{sinha2020multi,
  title={Multi-scale self-guided attention for medical image segmentation},
  author={Sinha, Ashish and Dolz, Jose},
  journal={IEEE Journal of Biomedical and Health Informatics},
  volume={25},
  number={1},
  pages={121--130},
  year={2021}
}

@inproceedings{pawlowski2018unsupervised,
  title={{Unsupervised Lesion Detection in Brain CT using Bayesian Convolutional Autoencoders}},
  author={Pawlowski, Nick and Lee, Matthew CH and Rajchl, Martin and McDonagh, Steven and Ferrante, Enzo and Kamnitsas, Konstantinos and Cooke, Sam and Stevenson, Susan and Khetani, Aneesh and Newman, Tom and others},
  booktitle={Medical Imaging with Deep Learning},
  year={2018}
}

@inproceedings{zimmerer2018context,
  title={{Context-encoding Variational Autoencoder for Unsupervised Anomaly Detection}},
  author={Zimmerer, David and Kohl, Simon AA and Petersen, Jens and Isensee, Fabian and Maier-Hein, Klaus H},
  booktitle={Medical Imaging with Deep Learning},
  year={2019}
}

@inproceedings{zimmerer2019unsupervised,
  title={Unsupervised anomaly localization using variational auto-encoders},
  author={Zimmerer, David and Isensee, Fabian and Petersen, Jens and Kohl, Simon and Maier-Hein, Klaus},
  booktitle={International Conference on Medical Image Computing and Computer-Assisted Intervention (MICCAI)},
  pages={289--297},
  year={2019}
}

@article{baur2021autoencoders,
  title={{Autoencoders for unsupervised anomaly segmentation in brain MR images: A comparative study}},
  author={Baur, Christoph and Denner, Stefan and Wiestler, Benedikt and Navab, Nassir and Albarqouni, Shadi},
  journal={Medical Image Analysis},
  volume={69},
  pages={101952},
  year={2021}
}

@inproceedings{bercea2023aes,
  title={{What Do AEs Learn? Challenging Common Assumptions in Unsupervised Anomaly Detection}},
  author={Bercea, Cosmin I and Rueckert, Daniel and Schnabel, Julia A},
  booktitle={International Conference on Medical Image Computing and Computer-Assisted Intervention (MICCAI)},
  pages={304--314},
  year={2023}
}

@inproceedings{Schlegl2017unsupervised,
  title={{Unsupervised Anomaly Detection with Generative Adversarial Networks to Guide Marker Discovery}},
  author={Schlegl, Thomas and Seeb{\"o}ck, Philipp and Waldstein, Sebastian M. and Schmidt-Erfurth, Ursula and Langs, Georg},
  booktitle={Information Processing in Medical Imaging},
  pages={146--157},
  year={2017},
}

@article{schlegl2019f,
  title={{f-AnoGAN: Fast unsupervised anomaly detection with generative adversarial networks}},
  author={Schlegl, Thomas and Seeb{\"o}ck, Philipp and Waldstein, Sebastian M and Langs, Georg and Schmidt-Erfurth, Ursula},
  journal={Medical Image Analysis},
  volume={54},
  pages={30--44},
  year={2019}
}

@inproceedings{baur2020steganomaly,
  title={{SteGANomaly: Inhibiting CycleGAN Steganography for Unsupervised Anomaly Detection in Brain MRI}},
  author={Baur, Christoph and Graf, Robert and Wiestler, Benedikt and Albarqouni, Shadi and Navab, Nassir},
  booktitle={International Conference on Medical Image Computing and Computer-Assisted Intervention (MICCAI)},
  pages={718--727},
  year={2020}
}

@article{ho2020denoising,
  title={Denoising diffusion probabilistic models},
  author={Ho, Jonathan and Jain, Ajay and Abbeel, Pieter},
  journal={Advances in Neural Information Processing Systems (NeurIPS)},
  volume={33},
  pages={6840--6851},
  year={2020}
}

@inproceedings{wolleb2022diffusion,
  title={Diffusion models for medical anomaly detection},
  author={Wolleb, Julia and Bieder, Florentin and Sandk{\"u}hler, Robin and Cattin, Philippe C},
  booktitle={International Conference on Medical Image Computing and Computer-Assisted Intervention (MICCAI)},
  pages={35--45},
  year={2022}
}

@inproceedings{wyatt2022anoddpm,
  title={{AnoDDPM: Anomaly detection with denoising diffusion probabilistic models using simplex noise}},
  author={Wyatt, Julian and Leach, Adam and Schmon, Sebastian M and Willcocks, Chris G},
  booktitle={Proceedings of the IEEE/CVF Conference on Computer Vision and Pattern Recognition Workshops},
  pages={650--656},
  year={2022}
}

@inproceedings{bercea2023mask,
  title={{Mask, Stitch, and Re-Sample: Enhancing Robustness and Generalizability in Anomaly Detection through Automatic Diffusion Models}},
  author={Cosmin I. Bercea and Michael Neumayr and Daniel Rueckert and Julia A Schnabel},
  booktitle={ICML 3rd Workshop on Interpretable Machine Learning in Healthcare (IMLH) },
  year={2023}
}

@inproceedings{behrendt2024patched,
  title={{Patched diffusion models for unsupervised anomaly detection in brain MRI}},
  author={Behrendt, Finn and Bhattacharya, Debayan and Kr{\"u}ger, Julia and Opfer, Roland and Schlaefer, Alexander},
  booktitle={Medical Imaging with Deep Learning},
  pages={1019--1032},
  year={2024}
}

@inproceedings{bercea2024diffusion,
  title={Diffusion models with implicit guidance for medical anomaly detection},
  author={Bercea, Cosmin I and Wiestler, Benedikt and Rueckert, Daniel and Schnabel, Julia A},
  booktitle={International Conference on Medical Image Computing and Computer-Assisted Intervention (MICCAI)},
  pages={211--220},
  year={2024}
}

@inproceedings{liang2024itermask,
  title={{IterMask2: Iterative unsupervised anomaly segmentation via spatial and frequency masking for brain lesions in MRI}},
  author={Liang, Ziyun and Guo, Xiaoqing and Noble, J Alison and Kamnitsas, Konstantinos},
  booktitle={International Conference on Medical Image Computing and Computer-Assisted Intervention (MICCAI)},
  pages={339--348},
  year={2024}
}

@inproceedings{behrendt2024leveraging,
  title={{Leveraging the mahalanobis distance to enhance unsupervised brain MRI anomaly detection}},
  author={Behrendt, Finn and Bhattacharya, Debayan and Mieling, Robin and Maack, Lennart and Kr{\"u}ger, Julia and Opfer, Roland and Schlaefer, Alexander},
  booktitle={International Conference on Medical Image Computing and Computer-Assisted Intervention (MICCAI)},
  pages={394--404},
  year={2024}
}

@article{bercea2025melba,
  title={{Denoising Diffusion Models for Anomaly Localization in Medical Images}},
  author={Bercea, Cosmin I and Cattin, Philippe C and Schnabel, Julia A and Wolleb, Julia},
  journal={Machine Learning for Biomedical Imaging},
  volume={3},
  pages={687--711},
  year={2025}
}

@inproceedings{lipman2022flow,
  title={Flow matching for generative modeling},
  author={Lipman, Yaron and Chen, Ricky TQ and Ben-Hamu, Heli and Nickel, Maximilian and Le, Matt},
  booktitle={International Conference on Learning Representations (ICLR)},
  year={2023}
}

@inproceedings{liu2023flow,
  title={{Flow Straight and Fast: Learning to Generate and Transfer Data with Rectified Flow}},
  author={Liu, Xingchao and Gong, Chengyue and Liu, Qiang},
  booktitle={International Conference on Learning Representations (ICLR)},
  year={2023}
}

@inproceedings{beizaee2025reflect,
  title={{REFLECT: Rectified Flows for Efficient Brain Anomaly Correction Transport}},
  author={Beizaee, Farzad and Hajimiri, Sina and Ben Ayed, Ismail and Lodygensky, Gregory and Desrosiers, Christian and Dolz, Jose},
  booktitle={International Conference on Medical Image Computing and Computer-Assisted Intervention (MICCAI)},
  pages={456--466},
  year={2025}
}

@inproceedings{martin2025pnp,
  title={{Pnp-flow: Plug-and-play image restoration with flow matching}},
  author={Martin, S{\'e}gol{\`e}ne and Gagneux, Anne and Hagemann, Paul and Steidl, Gabriele},
  booktitle={International Conference on Learning Representations (ICLR)},
  volume={2025},
  pages={45466--45492},
  year={2025}
}

@inproceedings{lugmayr2022repaint,
  title={{RePaint: Inpainting using denoising diffusion probabilistic models}},
  author={Lugmayr, Andreas and Danelljan, Martin and Romero, Andres and Yu, Fisher and Timofte, Radu and Van Gool, Luc},
  booktitle={IEEE/CVF Conference on Computer Vision and Pattern Recognition (CVPR)},
  pages={11461--11471},
  year={2022}
}

@inproceedings{hadzic2026restora,
  title={{Restora-Flow: Mask-Guided Image Restoration with Flow Matching}},
  author={Hadzic, Arnela and Thaler, Franz and Bogensperger, Lea and Joham, Simon Johannes and Urschler, Martin},
  booktitle={Proceedings of the IEEE/CVF Winter Conference on Applications of Computer Vision (WACV) },
  pages={4943--4952},
  year={2026}
}

@article{liew2022large,
  title={A large, curated, open-source stroke neuroimaging dataset to improve lesion segmentation algorithms},
  author={Liew, Sook-Lei and Lo, Bethany P and Donnelly, Miranda R and Zavaliangos-Petropulu, Artemis and Jeong, Jessica N and Barisano, Giuseppe and Hutton, Alexandre and Simon, Julia P and Juliano, Julia M and Suri, Anisha and others},
  journal={Scientific Data},
  volume={9},
  number={1},
  pages={320},
  year={2022}
}

@article{calabrese2022university,
  title={{The University of California San Francisco preoperative diffuse glioma MRI dataset}},
  author={Calabrese, Evan and Villanueva-Meyer, Javier E and Rudie, Jeffrey D and Rauschecker, Andreas M and Baid, Ujjwal and Bakas, Spyridon and Cha, Soonmee and Mongan, John T and Hess, Christopher P},
  journal={Radiology: Artificial Intelligence},
  volume={4},
  number={6},
  pages={e220058},
  year={2022}
}

@article{menze2014multimodal,
  title={{The multimodal brain tumor image segmentation benchmark (BRATS)}},
  author={Menze, Bjoern H and Jakab, Andras and Bauer, Stefan and Kalpathy-Cramer, Jayashree and Farahani, Keyvan and Kirby, Justin and Burren, Yuliya and Porz, Nicole and Slotboom, Johannes and Wiest, Roland and others},
  journal={IEEE Transactions on Medical Imaging},
  volume={34},
  number={10},
  pages={1993--2024},
  year={2015}
}

@article{bakas2017advancing,
  title={{Advancing the cancer genome atlas glioma MRI collections with expert segmentation labels and radiomic features}},
  author={Bakas, Spyridon and Akbari, Hamed and Sotiras, Aristeidis and Bilello, Michel and Rozycki, Martin and Kirby, Justin S and Freymann, John B and Farahani, Keyvan and Davatzikos, Christos},
  journal={Scientific Data},
  volume={4},
  number={1},
  pages={170117},
  year={2017}
}

@article{bakas2018identifying,
  title={{Identifying the best machine learning algorithms for brain tumor segmentation, progression assessment, and overall survival prediction in the BRATS challenge}},
  author={Bakas, Spyridon and Reyes, Mauricio and Jakab, Andras and Bauer, Stefan and Rempfler, Markus and Crimi, Alessandro and Shinohara, Russell Takeshi and Berger, Christoph and Ha, Sung Min and Rozycki, Martin and others},
  journal={arXiv preprint arXiv:1811.02629},
  year={2018}
}

@article{pielawski2020introducing,
  title={{Introducing Hann windows for reducing edge-effects in patch-based image segmentation}},
  author={Pielawski, Nicolas and W{\"a}hlby, Carolina},
  journal={PloS one},
  volume={15},
  number={3},
  pages={e0229839},
  year={2020}
}

@INPROCEEDINGS{Ronneberger2015-ih,
  title     = "{U-Net}: Convolutional Networks for Biomedical Image
               Segmentation",
  booktitle = "International Conference on Medical Image Computing and {Computer-Assisted} Intervention (MICCAI)",
  author    = "Ronneberger, Olaf and Fischer, Philipp and Brox, Thomas",
  pages     = "234--241",
  year      =  2015
}

@inproceedings{loshchilov2019adamw,
  title={Decoupled weight decay regularization},
  author={Loshchilov, Ilya and Hutter, Frank},
  booktitle={International Conference on Learning Representations (ICLR)},
  year={2019}
}

@article{guo2023encoder,
  title={Encoder-decoder contrast for unsupervised anomaly detection in medical images},
  author={Guo, Jia and Lu, Shuai and Jia, Lize and Zhang, Weihang and Li, Huiqi},
  journal={IEEE Transactions on Medical Imaging},
  volume={43},
  number={3},
  pages={1102--1112},
  year={2024},
}

@inproceedings{roth2022towards,
  title={Towards total recall in industrial anomaly detection},
  author={Roth, Karsten and Pemula, Latha and Zepeda, Joaquin and Sch{\"o}lkopf, Bernhard and Brox, Thomas and Gehler, Peter},
  booktitle={2022 IEEE/CVF Conference on Computer Vision and Pattern Recognition (CVPR)},
  pages={14298--14308},
  year={2022}
}

@inproceedings{caron2021emerging,
  title={Emerging properties in self-supervised vision transformers},
  author={Caron, Mathilde and Touvron, Hugo and Misra, Ishan and J{\'e}gou, Herv{\'e} and Mairal, Julien and Bojanowski, Piotr and Joulin, Armand},
  booktitle={2021 IEEE/CVF International Conference on Computer Vision (ICCV)},
  pages={9630--9640},
  year={2021}
}

@inproceedings{guo2025dinomaly,
  title={{Dinomaly: The less is more philosophy in multi-class unsupervised anomaly detection}},
  author={Guo, Jia and Lu, Shuai and Zhang, Weihang and Chen, Fang and Li, Huiqi and Liao, Hongen},
  booktitle={2025 IEEE/CVF Conference on Computer Vision and Pattern Recognition (CVPR)},
  pages={20405--20415},
  year={2025}
}

@inproceedings{schulthess2025anomaly,
  title={{Anomaly Detection by Clustering DINO Embeddings using a Dirichlet Process Mixture}},
  author={Schulthess, Nico and Konukoglu, Ender},
  booktitle={International Conference on Medical Image Computing and Computer-Assisted Intervention (MICCAI)},
  pages={46--56},
  year={2025}
}

@article{behrendt2026review,
  title={{A review of deep learning-based Unsupervised Anomaly Detection in brain MRI}},
  author={Behrendt, Finn and Bhattacharya, Debayan and Maack, Lennart and Kr{\"u}ger, Julia and Opfer, Roland and Schlaefer, Alexander},
  journal={Medical Image Analysis},
  volume = {112},
  pages={104076},
  year={2026}
}

@article{chen2020unsupervised,
  title={Unsupervised lesion detection via image restoration with a normative prior},
  author={Chen, Xiaoran and You, Suhang and Tezcan, Kerem Can and Konukoglu, Ender},
  journal={Medical Image Analysis},
  volume={64},
  pages={101713},
  year={2020}
}

@inproceedings{gu2024revisiting,
  title={Revisiting deep ensemble uncertainty for enhanced medical anomaly detection},
  author={Gu, Yi and Lin, Yi and Cheng, Kwang-Ting and Chen, Hao},
  booktitle={International Conference on Medical Image Computing and Computer-Assisted Intervention (MICCAI)},
  pages={520--530},
  year={2024}
}

@article{luo2025ura,
  title={{URA-Net: Uncertainty-integrated anomaly perception and restoration attention network for unsupervised anomaly detection}},
  author={Luo, Wei and Xing, Peng and Cao, Yunkang and Yao, Haiming and Shen, Weiming and Li, Zechao},
  journal={IEEE Transactions on Circuits and Systems for Video Technology},
  year={2025}
}
}

\end{document}